\documentclass{llncs}
\usepackage{graphicx}
\usepackage[labelformat=simple]{subcaption}
\usepackage{todonotes}
\usepackage{amssymb, amsmath}
\begin{document}
\title{Coevolution of Generative Adversarial Networks}
\author{Victor Costa \and Nuno Louren\c{c}o \and Penousal Machado}
\institute{CISUC, Department of Informatics Engineering \\ University of Coimbra, Coimbra, Portugal\\\email{\{vfc, naml, machado\}@dei.uc.pt}}
\maketitle

\begin{abstract}
Generative adversarial networks (GAN) became a hot topic, presenting impressive results in the field of computer vision. However, there are still open problems with the GAN model, such as the training stability and the hand-design of architectures. Neuroevolution is a technique that can be used to provide the automatic design of network architectures even in large search spaces as in deep neural networks. Therefore, this project proposes COEGAN, a model that combines neuroevolution and coevolution in the coordination of the GAN training algorithm. The proposal uses the adversarial characteristic between the generator and discriminator components to design an algorithm using coevolution techniques. Our proposal was evaluated in the MNIST dataset. The results suggest the improvement of the training stability and the automatic discovery of efficient network architectures for GANs. Our model also partially solves the mode collapse problem.
\end{abstract}

\section{Introduction}
Generative adversarial networks (GAN) \cite{NIPS2014_5423} gained relevance for presenting state-of-the-art results in generative models, mainly in the field of computer vision.
A GAN combines two deep neural networks, a discriminator and a generator, in an adversarial training where these networks are confronted in a zero-sum game between them.
The generator creates fake samples based on an input distribution.
The objective is to deceive the discriminator.
On the other hand, the discriminator learns to distinguish between these fake samples and the real input data.

Several works improving the GAN model were recently published, leveraging the quality of the results to impressive levels \cite{arjovsky2017wasserstein,karras2018progressive,mao2017least}.
However, there are still open problems, such as the vanishing gradient and the mode collapse problems, all of them leading to difficulties in the training procedure.
Although there are strategies to minimize the effect of those problems, they remain fundamentally unsolved \cite{gulrajani2017improved,salimans2016improved}.

Another issue, not related only to GANs but also to neural networks in general, is the necessity to define a network architecture previously.
In that case, the topology and hyperparameters are usually chosen empirically, thus spending human time in repetitive tasks such as fine-tuning.
However, there are approaches that can automatize the design of neural network architectures.

Neuroevolution is the application of evolutionary algorithms to provide the automatic design of neural networks.
In neuroevolution, both the network architecture (e.g., topology, hyperparameters and the optimization method) and the parameters (e.g., weights) used in each neuron can be evolved.
NeuroEvolution of Augmenting Topologies (NEAT) \cite{neat} is a well-known neuroevolution method that evolves the weights and topologies of neural networks.
In further experiments, NEAT was also successfully applied in a coevolution context \cite{stanley2004competitive}.
Moreover, DeepNEAT~\cite{miikkulainen2017evolving} was recently proposed to expand NEAT to larger search spaces, such as in deep neural networks.

Therefore, this project proposes a new model, called coevolutionary generative adversarial networks (COEGAN), to combine neuroevolution and coevolution in the coordination of the GAN training algorithm.
The evolutionary algorithm is based on DeepNEAT.
We extended and adapted this model to work on the context of GANs, making use of the competitive characteristic between the generator and discriminator to apply a coevolution model.
Hence, each subpopulation of generators and discriminators evolve following its own evolutionary path.
To validate our model, experiments were conducted using MNIST \cite{lecun1998mnist} as the input dataset for the discriminator component.
The results are the improvement of the training stability and the automatic discovery of efficient network topologies for GANs.

The remainder of this paper is organized as follows:
Section \ref{sec:background} introduces the concepts of GANs and evolutionary algorithms;
Section \ref{sec:proposal} presents our approach used to evolve GANs;
Section \ref{sec:experiments} displays the experimental results using this approach;
finally, Section \ref{sec:conclusions} presents conclusions and proposals for further enhancements.

\section{Background and Related Works} \label{sec:background}
This section reviews the concepts employed in this paper and presents works related to the proposed model.

\subsection{Evolutionary Algorithms} \label{sec:evolutionary_algorithms}
An evolutionary algorithm (EA) is a method inspired by biological evolution that aims to mimic the same evolutionary mechanism found in nature.
In EAs, the population is composed of individuals that represent possible solutions for a given problem, using a high-order abstraction to encode their characteristics \cite{sims1994evolving}.
The algorithm works by applying variation operators (e.g., mutation and crossover) to the population in order to search for better solutions.

\subsubsection{Neuroevolution.} \label{sec:neuroevolution}
Neuroevolution is the application of evolutionary algorithms in the evolution of neural networks.
This approach can be applied to weights, topology and hyperparameters of a neural network.
When used to generate a network topology, a substantial benefit is the automation of the architecture design and its parameters \cite{neat}.
Besides, not only the final architecture is important, but the intermediary models also give their contributions to the final model in form of the transference of their trained weights kept through generations \cite{neat}.
This automation is even more important with the rise of deep learning, which produces larger models and increases the search space \cite{assunccao2018evolving,miikkulainen2017evolving}.
However, large search spaces are also a challenge for neuroevolution.
These methods have a high computational complexity that may turn their application unfeasible.

\subsubsection{Coevolution.} \label{sec:coevolution}
The simultaneous evolution of at least two distinct species is called coevolution \cite{hillis1990co,rawal2010constructing}.
There are two types of coevolution algorithms: cooperative and competitive.
In cooperative coevolution, individuals of different species cooperate in the search for efficient solutions, and the fitness function of each species are designed to reward this cooperation \cite{garcia2003covnet,garcia2005cooperative,gomez2008accelerated}.
In competitive coevolution, individuals of different species are competing between them.
Consequently, their fitness function directly represents this competition in a way that scores between species are inversely related \cite{stanley2004competitive,sims1994evolving,rawal2010constructing}.

\subsection{Generative Adversarial Networks} \label{sec:gan}
Generative Adversarial Networks (GAN), proposed in \cite{NIPS2014_5423}, is an adversarial model that became relevant for the performance achieved in generative tasks.
A GAN combines two deep neural networks: a discriminator $D$ and a generator $G$.
The generator $G$ receives a noise as input and outputs a fake sample, attempting to capture the data distribution used as input for $D$.
The discriminator $D$ receives the real data and fake samples as input, learning to distinguish between them.
These components are trained simultaneously as adversaries, creating strong generative and discriminative components.

The loss function for the discriminator is defined by:

\begin{equation}
J^{(D)}(D,G) = -\mathbb{E}_{x \sim p_{data}}[\log D(x)] - \mathbb{E}_{z \sim p_z}[\log(1 - D(G(z)))].
\label{eq:discriminator}
\end{equation}

The loss function for the generator (non-saturating version proposed in \cite{NIPS2014_5423}) is defined by:

\begin{equation}
J^{(G)}(G) = - \mathbb{E}_{z \sim p_z}[\log(D(G(z)))].
\label{eq:generator}
\end{equation}

In Eq. \ref{eq:discriminator}, $p_{data}$ represents the dataset used as input to the discriminator.
In Eq. \ref{eq:discriminator} and Eq. \ref{eq:generator}, $z$, $p_z$, $G$ and $D$ represent the noise sample (used as input to the generator), the noise distribution, the generator and the discriminator, respectively.

Besides, several variations of the loss function were proposed to improve the GAN model, such as in WGAN \cite{arjovsky2017wasserstein} and LSGAN \cite{mao2017least}.
These variations were studied in \cite{lucic2017gans} in order to access the superiority in respect to the original GAN proposal.
The study founds no empirical evidence that these variations are superior to the original GAN mode.

There are two common problems regarding training stability in GANs: vanishing gradient and mode collapse.
The vanishing gradient occurs when the discriminator $D$ became perfect and do not commit mistakes anymore.
Hence, the loss function is zeroed, the gradient does not flow through the neural network of the generator, and the GAN progress stagnates.
In mode collapse, the generator captures only a small portion of the dataset distribution provided as input to the discriminator.
This is not desirable once we want to reproduce the whole distribution of the data.
Recently, several approaches tried to minimize those problems, but they remain unsolved \cite{salimans2016improved,gulrajani2017improved}.

There are other models extending the original GAN proposal that modify not only the loss function but also aspects of the architecture.
The method described by \cite{karras2018progressive} uses a simple strategy to evolve a GAN during the training procedure.
The main idea is to grow the model progressively, increasing layers in both discriminator and generator.
This mechanism will make the model more complex while the training proceeds, increasing the resolution of images at each phase.
However, these layers added progressively are preconfigured, i.e., they are hand-designed and not produced by a stochastic procedure.
Thus, the model evolves in a preconfigured way during the training procedure, but the method used in \cite{karras2018progressive} do not use an evolutionary algorithm in this process.
Therefore, we can consider this predefined progression as a first step towards the evolution of generative adversarial models.

A very recent model proposes the use of evolutionary algorithms in GANs \cite{wang2018evolutionary}.
Their approach used a simple model for the evolution, using a mutation operator that can change only the loss function of the individuals.
Our proposal differs from them by modeling the GAN as a coevolution problem.
Besides, in our case, the evolution occurs in the network architecture, and the loss function is the same during the whole method.
Nevertheless, a further proposal can incorporate those ideas to evaluate the benefits.

\section{Coevolution of Generative Adversarial Networks} \label{sec:proposal}
We propose a new model called coevolutionary generative adversarial networks (COEGAN).
This model combines neuroevolution and coevolution in the coordination of the GAN training algorithm.
Our approach is based on DeepNEAT \cite{miikkulainen2017evolving}, that was extended and adapted to the context of GANs.

In COEGAN, the genome is represented as an array of genes, which are directly mapped into a phenotype consisting of the sequence of layers in a deep neural network.
Each gene represents a linear, convolution or transpose convolution layer.
Moreover, each gene also has an activation function, chosen from the following set: ReLU, LeakyReLU, ELU, Sigmoid and Tanh.
From the specific parameters of each type of gene, convolution and transpose convolution layers only have the number of output channels as a random parameter.
The stride and kernel size are fixed as $2$ and $3$, respectively.
The number of input channels is calculated dynamically, based on the previous layer.
Similarly, the linear layer only has the number of output features as the random parameter.
The number of input features is calculated based on the previous layer.
Therefore, only the activation function, output features and output channels are subject to the variation operations.

\begin{figure}
	\centering
	\begin{subfigure}[t]{.5\textwidth}
		\centering
		\includegraphics[width=\textwidth]{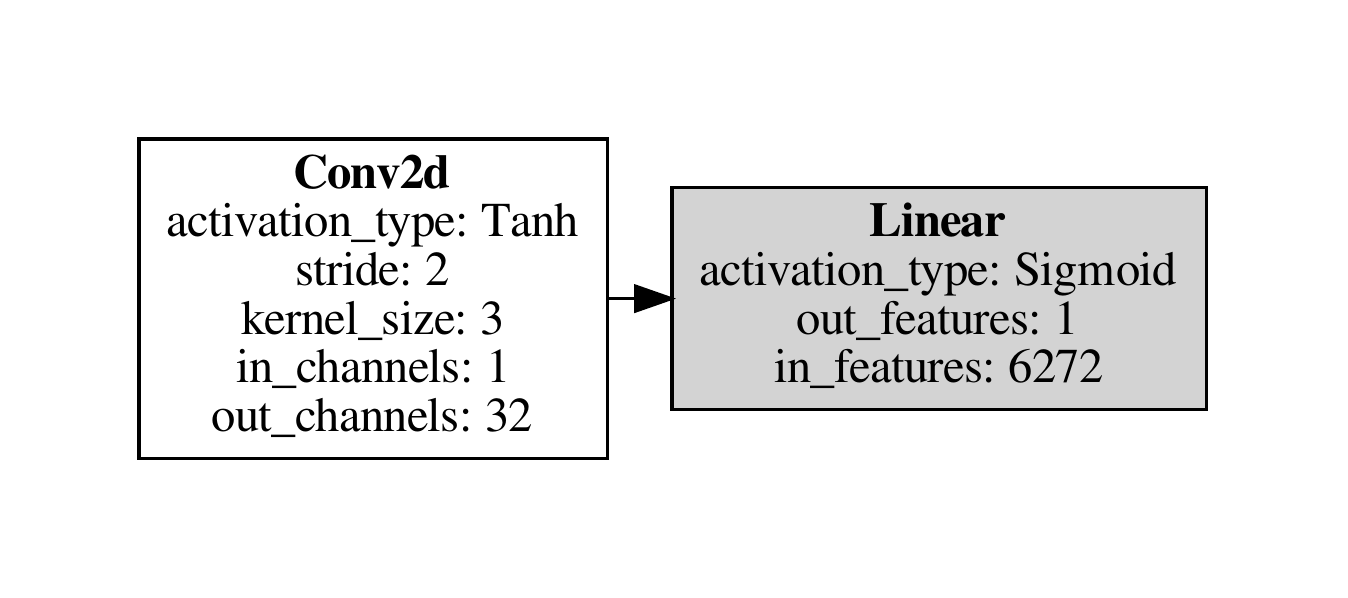}
		\caption{A genotype of a discriminator}
		\label{fig:discriminator_genotype}
	\end{subfigure}%
	\begin{subfigure}[t]{.5\textwidth}
		\centering
		\includegraphics[width=\textwidth]{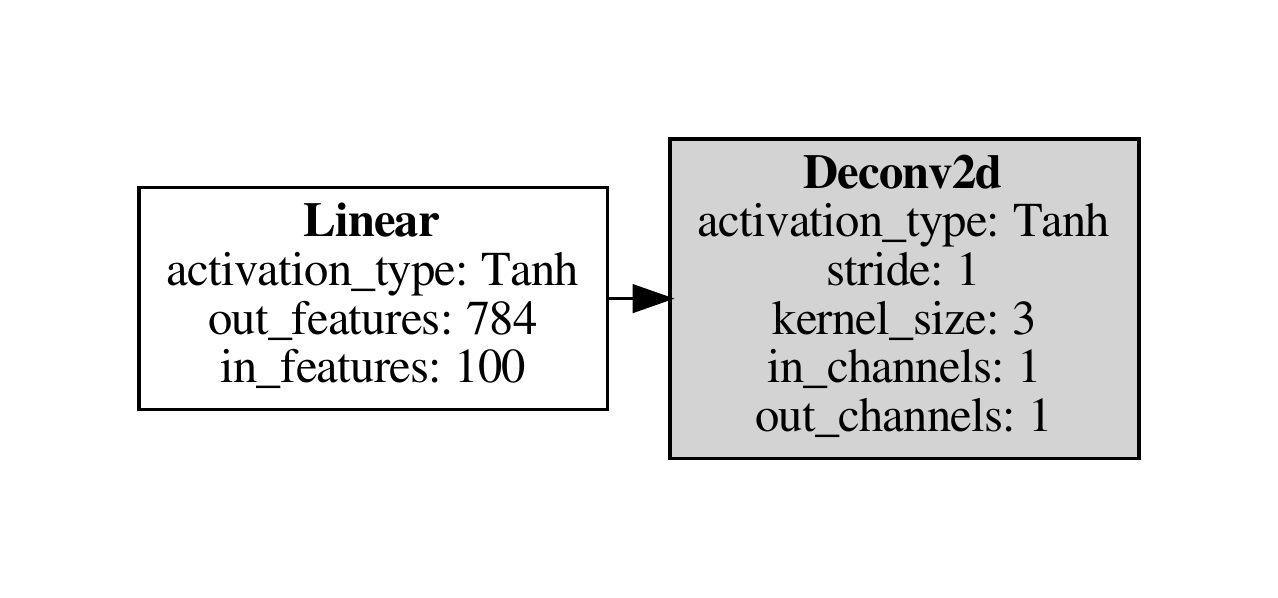}
		\caption{A genotype of a generator}
		\label{fig:generator_genotype}
	\end{subfigure}
	\caption{Example of genotypes of a generator and a discriminator.}
	\label{fig:mnist_genotypes}
\end{figure}

Figures \ref{fig:discriminator_genotype} and \ref{fig:generator_genotype} are examples of a discriminator and a generator genotype, respectively.
The discriminator genotype is composed of a convolutional section and followed by a linear section (fully connected layers).
As in the original GAN approach, the output of discriminators is the probability of the input sample be a real sample drawn from the dataset.
Similarly, the generator genotype is composed of a linear section and followed by a transpose convolutional section.
The output of the generator is a fake sample, with the same characteristics (i.e., dimension and channels) of a real sample.

The overall population is composed of two separated subpopulations: a population of generators, where each $G_i$ represents a generator component in a GAN; and a population of discriminators where each $D_j$ represents a discriminator in a GAN.
Furthermore, a speciation mechanism based on the original NEAT proposal is applied to promote innovation in each subpopulation.
The speciation mechanism divides the population into species based on a similarity function (used to group similar individuals).
Thus, the innovation, represented by the addition of new genes into a genome, causes the creation of new species in order to fit the individuals containing these new genes.
Therefore, these new individuals have the chance to survive through generations and reach the performance of older individuals in the population.

The parameters of the layers in the phenotype (e.g., weights and bias) will be trained by the gradient descent method and will not be part of the evolution.
The number of parameters to be optimized are too large and evolving them will increase the computational complexity.
Therefore, for now, we are interested only in the evolution of the network topology.
We plan to develop a hybrid approach that evolves the weight when the gradient descent training stagnates for a number of generations.

\subsection{Fitness} \label{sec:fitness}

For discriminators, the fitness is based on the loss obtained from the regular GAN training method, i.e., the fitness is equivalent to Eq. \ref{eq:discriminator}.
We have tried to use the same approach for the generator.
However, preliminary experiments evidenced that the loss does not represent a good measure for quality in this case.
The loss for generators, represented by Eq. \ref{eq:generator}, is unstable during the GAN training, making it not suitable to be used as fitness in an evolutionary algorithm.

Thus, we selected the Fr\'{e}chet Inception Distance (FID) \cite{heusel2017gans} as the fitness for generators.
FID is the state-of-the-art metric to compare the generative components of GANs and outperforms other metrics, such as the Inception Score \cite{salimans2016improved}, with respect to diversity and quality \cite{lucic2017gans}.
In FID, an Inception Net~\cite{szegedy2016rethinking} (trained on ImageNet~\cite{russakovsky2015imagenet}) is used to transform images to the feature space (given by a hidden layer of the network). This feature space is interpreted as a continuous multivariate Gaussian.
So, the mean and covariance of two Gaussians are estimated using real and fake samples.
The Fr\'{e}chet distance between these Gaussians is given by:

\begin{equation}
FID(x,g) = ||\mu_x - \mu_g||_2^2 + Tr(\varSigma_x + \varSigma_g - 2(\varSigma_x\varSigma_g)^{1/2}).
\label{eq:fid}
\end{equation}

In Eq. \ref{eq:fid}, $\mu_x$, $\varSigma_x$, $\mu_g$, and $\varSigma_g$ represent the mean and covariance estimated for the real dataset and for fake samples, respectively.

\subsection{Variation Operators} \label{sec:variation_operators}
\begin{figure}[ht]
	\includegraphics[width=\textwidth]{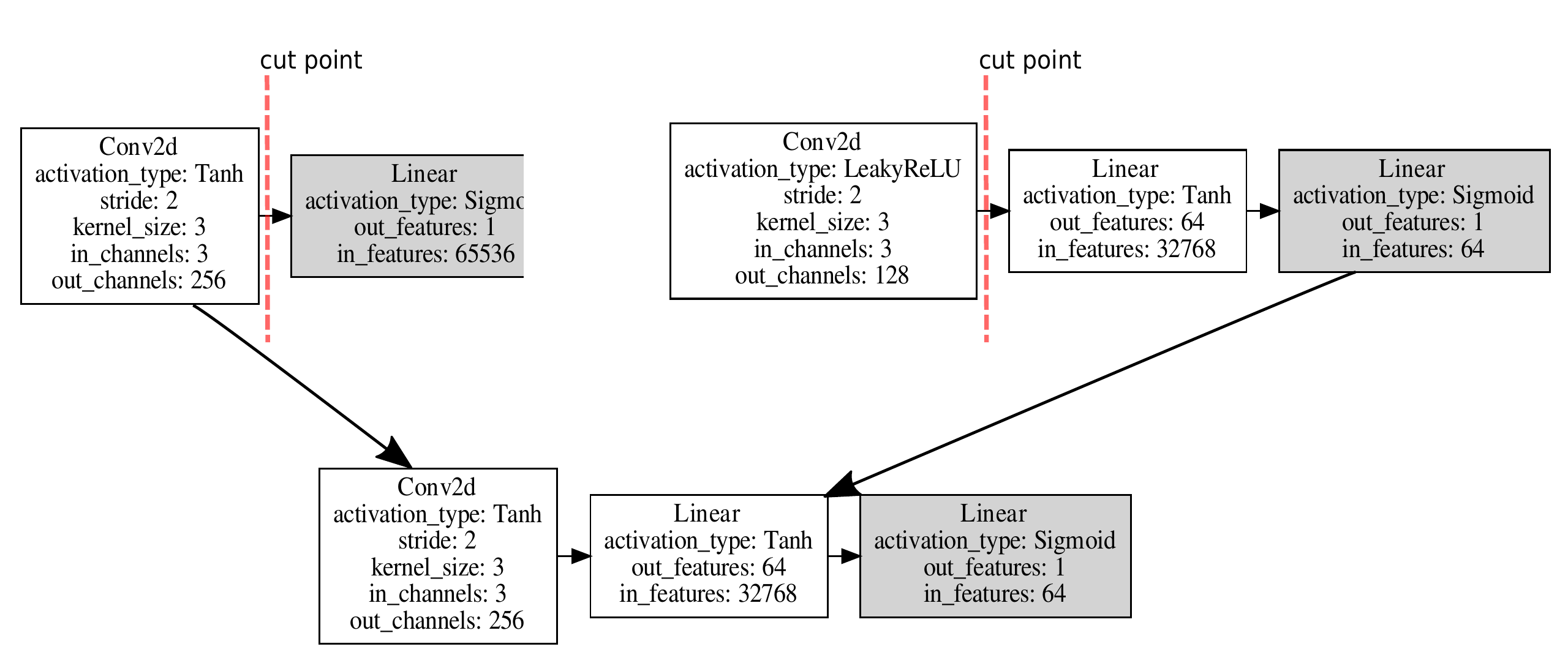}
	\caption{Example of crossover between discriminators.}
	\label{fig:discriminator_crossover}
\end{figure}

Initially, we have used two types of variation operators to breed new individuals: mutation and crossover.
The crossover process uses the transition between convolutional layers and linear (fully connected) layers as the cut point.
Figure \ref{fig:discriminator_crossover} represents an example of this process.
However, preliminary tests evidenced that crossover decreases the performance of the system.
We expect to conduct further experiments with other crossover variations to assess the contribution of this operator in our model.

The mutation process is composed of three main operations: add a new layer, remove a layer, and change an existing layer.
In the addition operation, a new layer is randomly drawn from the set of possible layers.
For discriminators, the available layers are linear and convolution.
For generators, the available layers are linear and transpose convolution (also called deconvolution).
The remove operation chooses an existing layer and excludes it from the genotype.

The change operation modifies the attributes and the activation function of an existing layer.
The activation function is randomly chosen from the set of possibilities.
Other specific attributes can be changed depending on the type of the layer.
The number of output features and the number of output channels are mutated for the linear and convolution layers, respectively.
The mutation of these attributes follows a uniform distribution, with a predefined range limiting the possible values.

It is important to note that if the new gene is compatible with its parent, the parameters (weights and bias) are copied.
So, the new individual will carry the training information from the previous generation.
This simulates the transfer learning technique commonly used to train deep neural networks.
However, when the attributes of a linear or convolution layer change, the trained parameters are lost.
This happens because the setup of weights changes, becoming incompatible with the new layer.
Consequently, the weights are not transferred from the parents and the layer will be trained from the beginning.

\subsection{Competition Between Generators and Discriminators}
In the evaluation step of the evolutionary algorithm, discriminators and generators must be paired to calculate the fitness for each individual in the population.
There are several approaches to pair individuals in competitive coevolution, e.g. \textit{all vs. all}, \textit{random} and \textit{all vs. best} \cite{sims1994evolving}.

\begin{figure}
	\centering
	\includegraphics[width=0.3\textwidth]{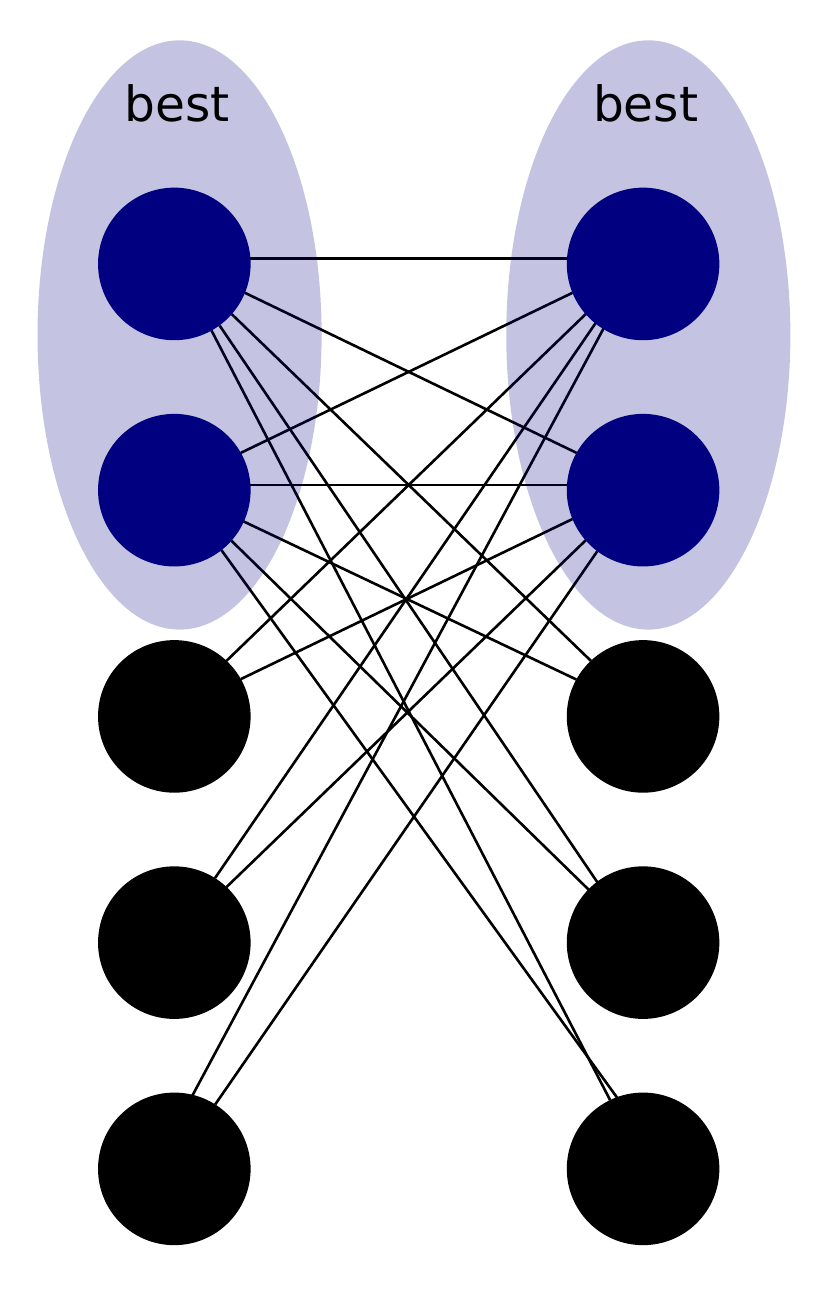}
	\caption{Representation of the all vs. best competition pattern with $k=2$.}
	\label{fig:all_vs_best}
\end{figure}

As we want to train the GAN to avoid problems such as the mode collapse and vanishing gradient, we consider the use of the \textit{all vs. best} strategy here.
However, we select $k$ individuals rather than only one to promote the diversity in the GAN training.
We pair each generator with $k$ best discriminators from the previous generation and, similarly, each discriminator with $k$ best generators.
Figure \ref{fig:all_vs_best} represents an example of this approach with $k=2$.
For the first generation, we assume a random approach, i.e., $k$ random individuals are selected to be paired in the initial evaluation.

The \textit{all vs. all} strategy would also be interesting for our model as it will improve the variability of the environment for both discriminators and generators during the training.
However, the trade-off is the time to execute this approach.
In \textit{all vs. all}, each discriminator is paired with each generator, resulting in many competitions.

\subsection{Selection} \label{sec:selection}
For the selection phase, we used a strategy based on NEAT~\cite{neat}.
As in NEAT, we divided the population of generators and discriminators into subpopulations, following a speciation strategy similar to that used in NEAT.
Each species contains individuals with similar network structures.
For this, we define the similarity between individuals based on the parameters of each gene composing the genome.
Different of NEAT, we do not use the weights of each layer to differentiate between individuals.
Therefore, we calculate the distance $\delta$ between two genomes $i$ and $j$ as the number of genes that exist only in $i$ or $j$.
Each species inside the populations of generators and discriminators are clustered based on a threshold $\delta_t$.
This threshold is calculated in order to suit the desired number of species.
The number of species is a parameter previously chosen. 

The selection occurs inside each species.
The number of individuals selected inside each species are proportional to the average fitness of the individuals belonging to it.
Given the number of individuals to keep in a species, a tournament between $k_t$ individuals is applied to finally select the individuals to breed and compose the next generation.

\section{Experiments} \label{sec:experiments}
In this section, we will evaluate the performance of our method on the MNIST dataset.
Normally, the network would be training for several epochs using the whole dataset in the procedure.
As this would be an intensive computational task, we will use only a subset of the dataset per generation.
This strategy, combined with the transfer of parameters between generations, was sufficient to produce an evolutionary pressure towards efficient solutions and to promote the GAN convergence.

There is no consensus on the metric to represent the quality of samples generated by generative models.
However, the Fr\'{e}chet Inception Distance (FID) was proved to be the best metric when comparing the quality of samples generated by GANs \cite{lucic2017gans}.
Therefore, we used the FID score, the same metric used as fitness for generators, to compare our results with the state of the art.

\subsection{Experimental setup}

Table \ref{table:setup} describes the parameters used in all experiments reported in this paper.

\begin{table}
	\caption{Experimental parameters.}
	\begin{center}\begin{tabular}{c|c}
		\textbf{Evolutionary Parameters} & \textbf{Value} \\
		\hline
		Number of generations & 100 \\
		Population size (generators) & 20 \\
		Population size (discriminators) & 20 \\
		Crossover rate & 0\% \\
		Add Layer rate & 30\% \\
		Remove Layer rate & 10\% \\
		Change Layer rate & 10\% \\
		Output features range & [32, 1024] \\
		Output channels range & [16, 128] \\
		$k$ (all vs. best) & 3 \\
		Tournament $k_t$ & 2 \\
		FID samples & 1000 \\
		Genome Limit & 4 \\
		Species & 4 \\
		\textbf{GAN Parameters} & \textbf{Value} \\
		\hline
		Batch size & 64 \\
		Batches per generation & 20 \\
		Optimizer & RMSProp \\
		Learning rate & 0.001
	\end{tabular}\end{center}
	\label{table:setup}
\end{table}

For evolutionary parameters, we chose to execute our experiments for 100 generations.
After this number of generations, the fitness stagnates and we expect no improvement of the results.
We used 20 individuals for the population of both generators and discriminators.
A larger population will probably achieve better results, but the computational cost would be too large.
The size of the genome was limited to four layers, also to reduce the computational cost.
The number of species used was four, permitting an average of five individuals per species in each subpopulation (generators and discriminators).
We empirically defined a probability of 30\%, 10\% and 10\% for the add, remove and change mutations, respectively.
As stated before, crossover was not used in the experiments reported in this section.

For the GAN parameters, we choose $64$ as batch size, running $20$ batches per generation.
This amounts to $1280$ samples per generation to train discriminators.
The optimizer used in the training method was RMSProp~\cite{tieleman2012lecture}.
We have also conducted preliminary experiments with Adam~\cite{kingma2015adam}, but the best results were achieved with RMSProp.

The MNIST dataset was used and we executed each experiment $10$ times to achieve the results within a confidence interval of 95\%.

\subsection{Results}

\begin{figure}[ht]
	\centering
	\begin{subfigure}[t]{.5\textwidth}
		\centering
		\includegraphics[width=\textwidth]{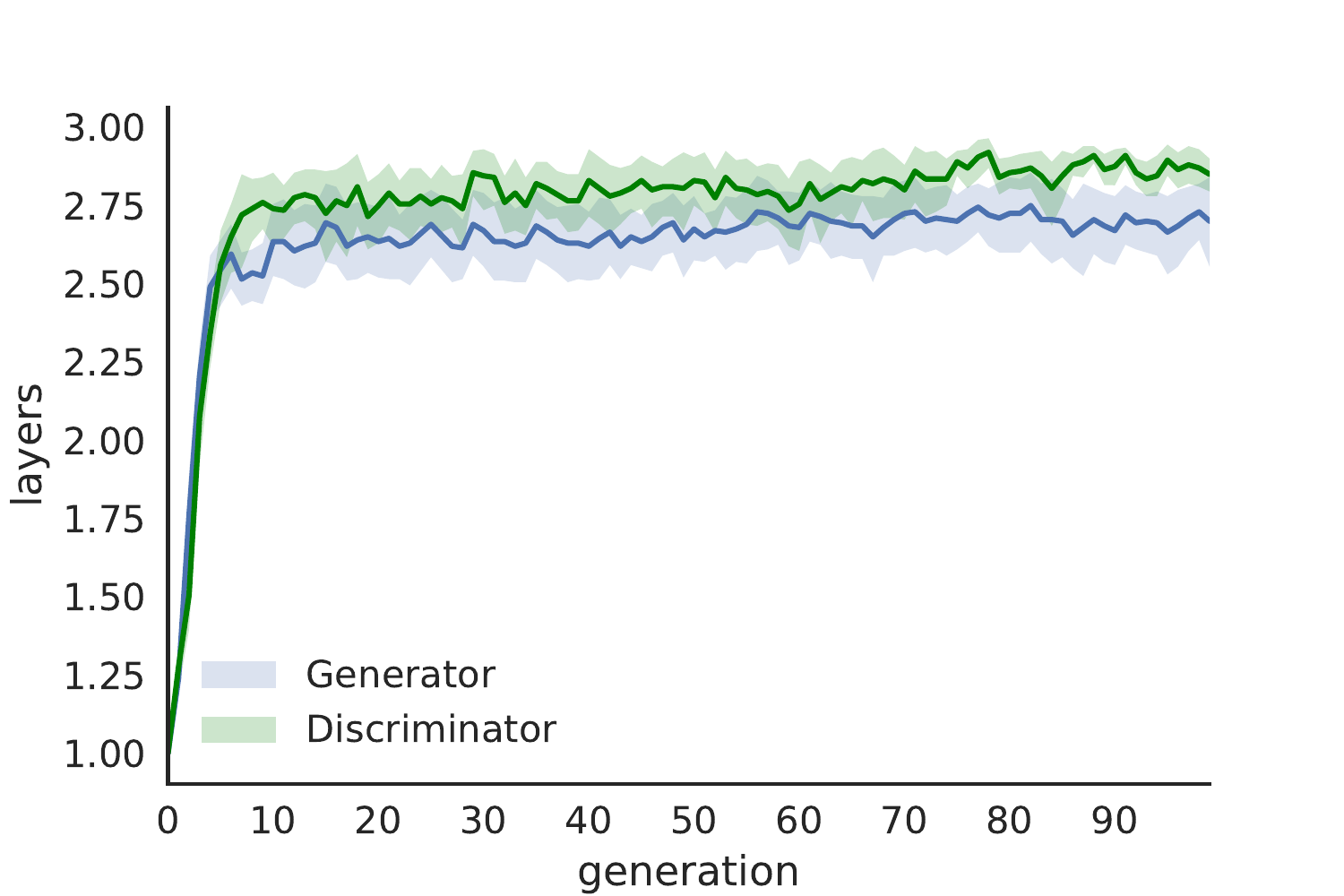}
		\caption{Layers per generation}
		\label{fig:mnist_layers}
	\end{subfigure}%
	\begin{subfigure}[t]{.5\textwidth}
		\centering
		\includegraphics[width=\textwidth]{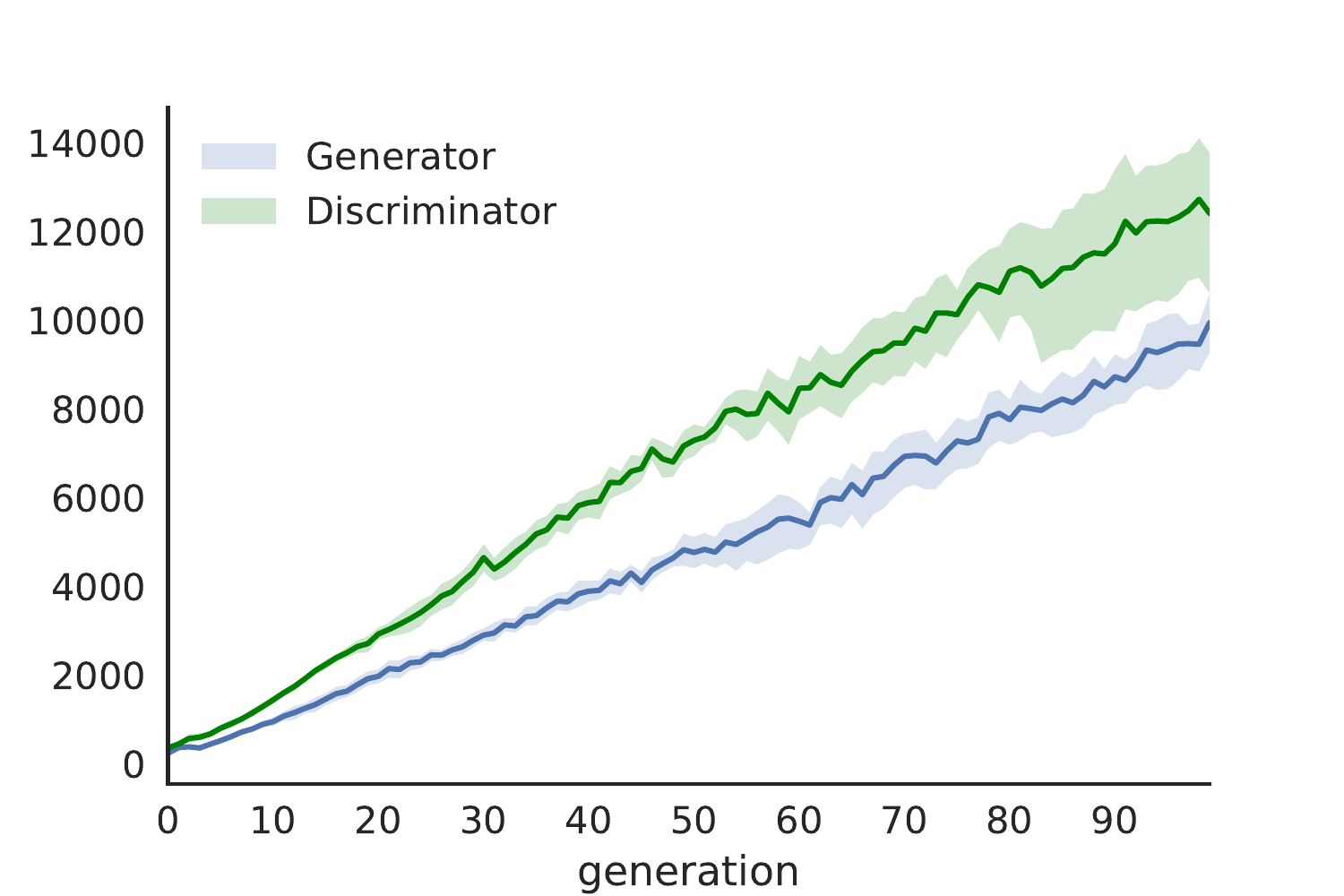}
		\caption{Reused genes}
		\label{fig:mnist_genes}
	\end{subfigure}
	\caption{Progression of layers and the reuse of parameters with a 95\% confidence interval}
	\label{fig:mnist_layers_genes}
\end{figure}

Figure \ref{fig:mnist_layers_genes} shows the progression of the network through generations.
We can see in \ref{fig:mnist_layers} the average number of layers in the population of generators and discriminators.
Because we have limited the genotype to a maximum of four genes, the number of layers rapidly saturates.
This is an indication of premature optimization.
We can overcome this issue by either increasing the limit or decreasing the growth rate (i.e., reduce the mutation probability).
In our tests with crossover activated this problem became even more evident.
Figure \ref{fig:mnist_genes} shows the number of genes with the parameters reused in each generation.
The linear growth in the amount of reused genes is evidence of the transference mechanism explained in Section \ref{sec:variation_operators}.
Because we use a strategy similar to transfer learning to keep the trained parameters, this reuse is important to pass the trained weights trough generations.

\begin{figure}[ht]
	\centering
	\begin{subfigure}[t]{.5\textwidth}
		\centering
		\includegraphics[width=\textwidth]{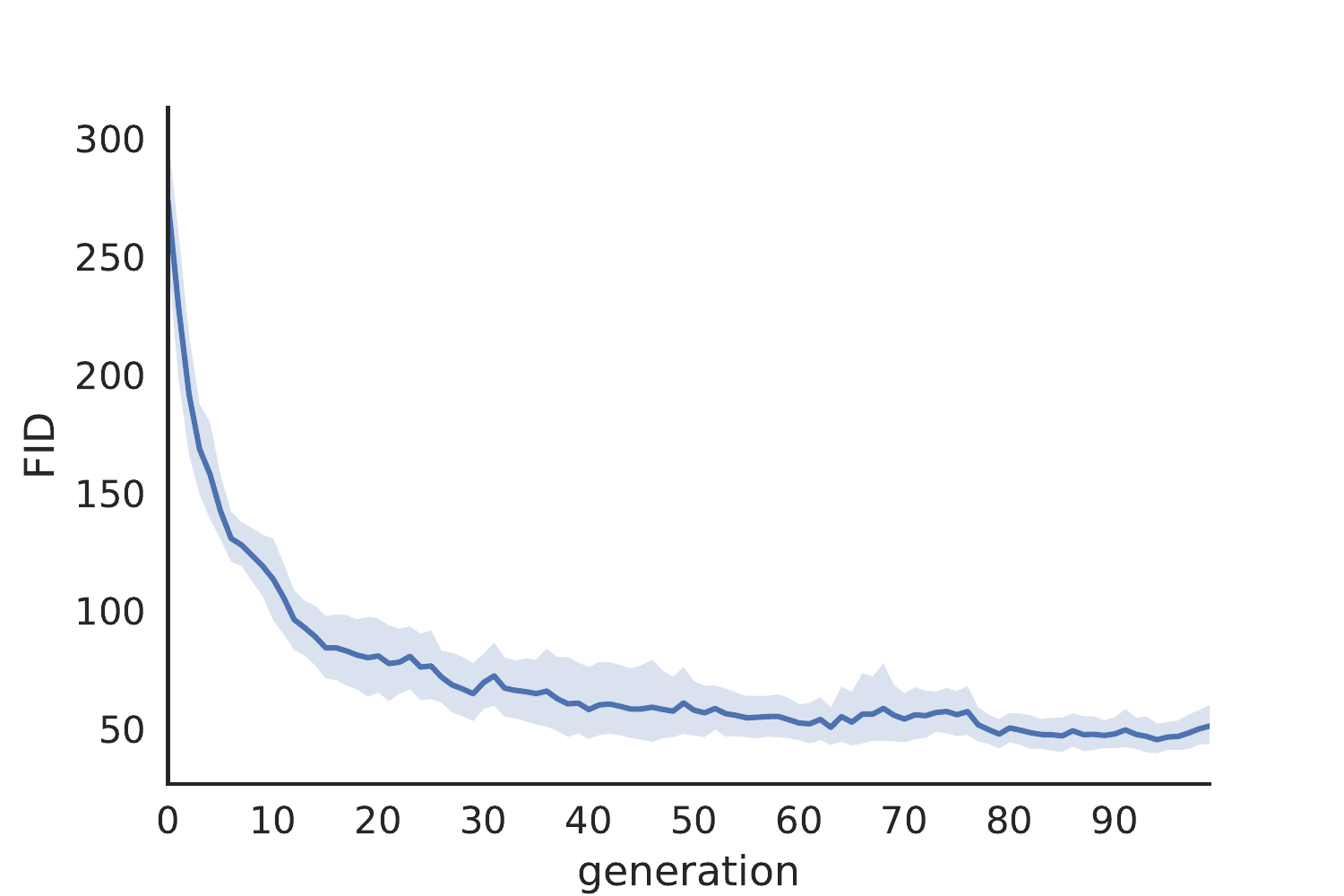}
		\caption{Fitness for generators}
		\label{fig:mnist_fitness_g}
	\end{subfigure}%
	\begin{subfigure}[t]{.5\textwidth}
		\centering
		\includegraphics[width=\textwidth]{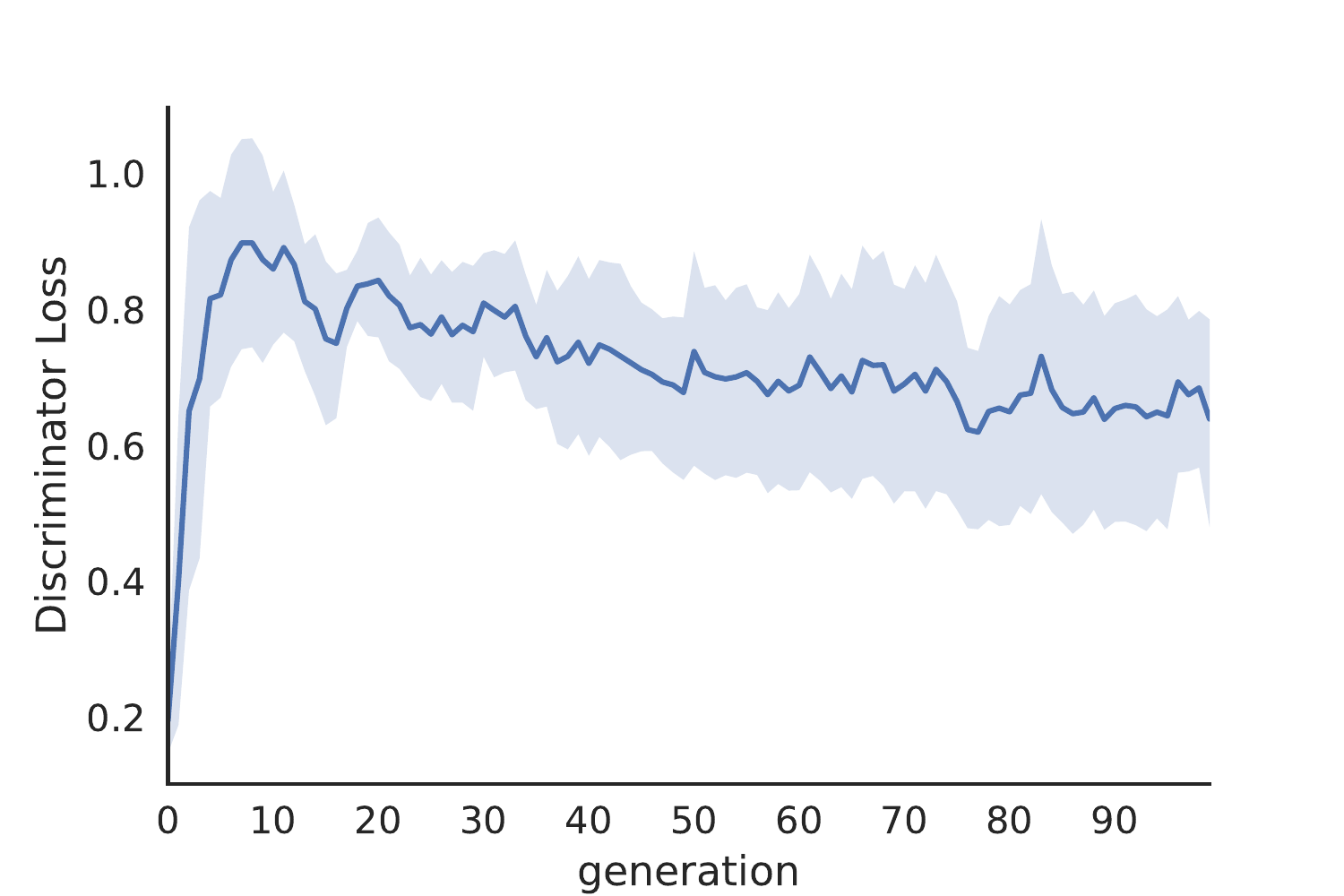}
		\caption{Fitness for discriminators}
		\label{fig:mnist_fitness_d}
	\end{subfigure}
	\caption{Fitness for discriminators and generators with a 95\% confidence interval}
	\label{fig:mnist_fitness}
\end{figure}

Figure \ref{fig:mnist_fitness} shows the progression of the fitness for best generators and discriminators.
In Figure \ref{fig:mnist_fitness_g}, we can see the fitness for generators reducing through generations with reduced noise.
Hence, the chosen fitness, i.e., the FID score, is evidenced as suitable to be used in our evolutionary algorithm.
For discriminators (Figure \ref{fig:mnist_fitness_d}), we can see much more noise, which can harm the selection process in the evolutionary algorithm.
This suggests that the choice for the discriminator fitness could be improved.

In the final generation, the mean FID was $49.2$, with a standard deviation of $10.5$.
The high standard deviation clearly shows that we need to increase the number of runs in order to get a more representative value for the FID.
Besides that, we can see that this score is much worse than state-of-the-art results.
For example, the FID for MNIST was reported in \cite{lucic2017gans} as $6.7$, with a standard deviation of $0.3$.
However, our results showed that the model did not collapse into a single point from the input distribution, which is a common problem in GANs.

\begin{figure}[ht]
	\centering
	\includegraphics[width=\textwidth]{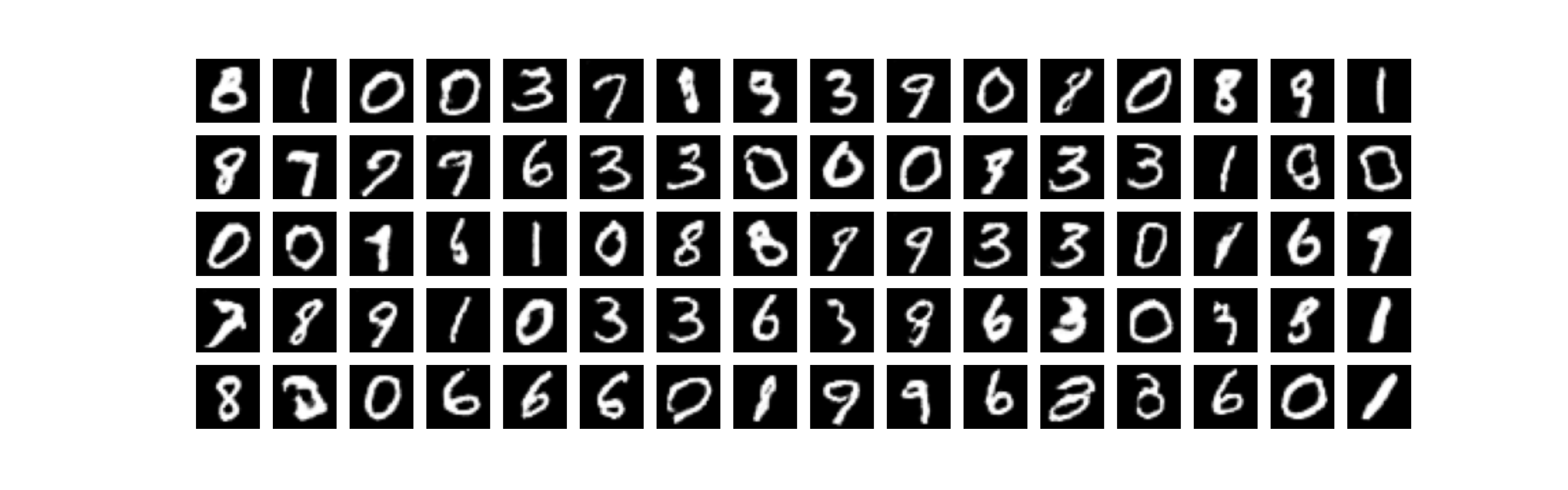}
	\caption{Samples created by a generator after the evolutionary algorithm.}
	\label{fig:mnist_samples}
\end{figure}

Figure \ref{fig:mnist_samples} represents samples generated in one execution, evidencing that our model does not collapse into a single point of the input distribution.
We can see this behavior occurring in all executions, which leave us to conclude that our model solves, at least partially, the mode collapse problem.
Moreover, all executions reached convergence with equilibrium between the discriminator and the generator, and the vanishing gradient never occurs.
This evidences that our proposal brings stability to the training procedure of GANs.
Furthermore, our experiments were restricted to a maximum of four layers.
A similar number of layers was used by \cite{lucic2017gans}.
This evidences that our method does not outperform architectures designed by hand when taking into account only the FID score.

For some executions, the generator captured only a subset of the distribution, which is a form of the mode collapse problem.
See in Figure \ref{fig:mnist_samples} examples of images created by a generator after the whole evolutionary algorithm.
Note that only half of the digits are represented in these samples.

\begin{figure}[ht]
	\centering
	\begin{subfigure}[t]{.28\textwidth}
		\centering
		\includegraphics[width=\textwidth]{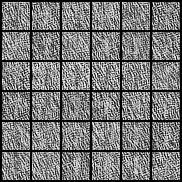}
		\caption{Generation 1}
		\label{fig:gen1}
	\end{subfigure}%
	\begin{subfigure}[t]{.28\textwidth}
		\centering
		\includegraphics[width=\textwidth]{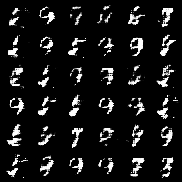}
		\caption{Generation 5}
		\label{fig:gen5}
	\end{subfigure}
	\begin{subfigure}[t]{.28\textwidth}
		\centering
		\includegraphics[width=\textwidth]{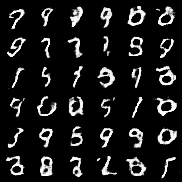}
		\caption{Generation 10}
		\label{fig:gen10}
	\end{subfigure}
	\begin{subfigure}[t]{.28\textwidth}
		\centering
		\includegraphics[width=\textwidth]{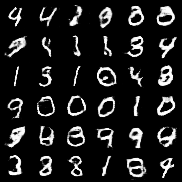}
		\caption{Generation 20}
		\label{fig:gen20}
	\end{subfigure}
	\begin{subfigure}[t]{.28\textwidth}
		\centering
		\includegraphics[width=\textwidth]{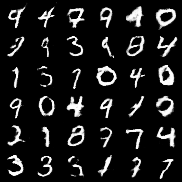}
		\caption{Generation 30}
		\label{fig:gen30}
	\end{subfigure}
	\begin{subfigure}[t]{.28\textwidth}
		\centering
		\includegraphics[width=\textwidth]{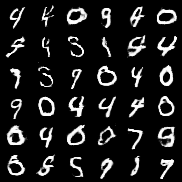}
		\caption{Generation 40}
		\label{fig:gen40}
	\end{subfigure}
	\begin{subfigure}[t]{.28\textwidth}
		\centering
		\includegraphics[width=\textwidth]{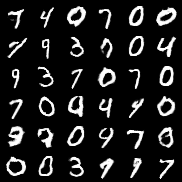}
		\caption{Generation 50}
		\label{fig:gen50}
	\end{subfigure}
	\begin{subfigure}[t]{.28\textwidth}
		\centering
		\includegraphics[width=\textwidth]{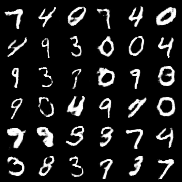}
		\caption{Generation 70}
		\label{fig:gen70}
	\end{subfigure}
	\begin{subfigure}[t]{.28\textwidth}
		\centering
		\includegraphics[width=\textwidth]{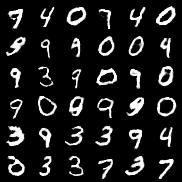}
		\caption{Generation 100}
		\label{fig:gen100}
	\end{subfigure}
	\caption{The progression of samples created by the best generator in generations \subref{fig:gen1} 1, \subref{fig:gen5} 5, \subref{fig:gen10} 10, \subref{fig:gen20} 20, \subref{fig:gen30} 30, \subref{fig:gen40} 40, \subref{fig:gen50} 50, \subref{fig:gen70} 70, and \subref{fig:gen100} 100.}
	\label{fig:gen_progression}
\end{figure}

Figure \ref{fig:gen_progression} contains generated samples selected to represent the progression of the generator during the evolutionary algorithm.
We can see in the first generation only noisy samples, without any structure resembling a digit.
After 5 generations (Figure \ref{fig:gen5}) we can see some structure emerging to form the digits.
From the generation 10 onwards we can start to distinguish between the digits, with a progressive improvement of the quality.

\begin{figure}[ht]
	\centering
	\begin{subfigure}[t]{\textwidth}
		\centering
		\includegraphics[width=\textwidth]{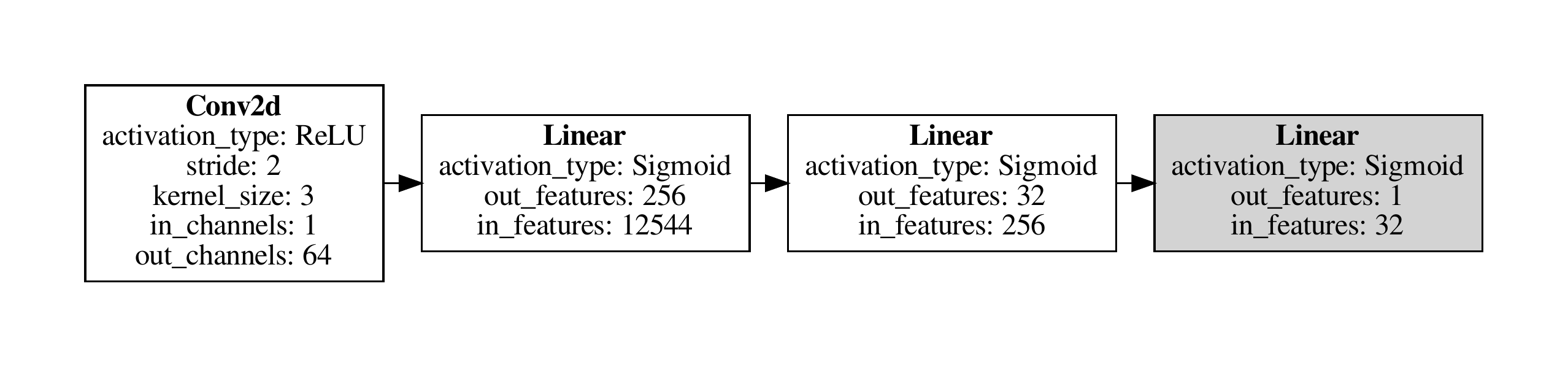}
		\caption{Genotype of a discriminator}
		\label{fig:mnist_discriminator}
	\end{subfigure}
	\begin{subfigure}[t]{\textwidth}
		\centering
		\includegraphics[width=\textwidth]{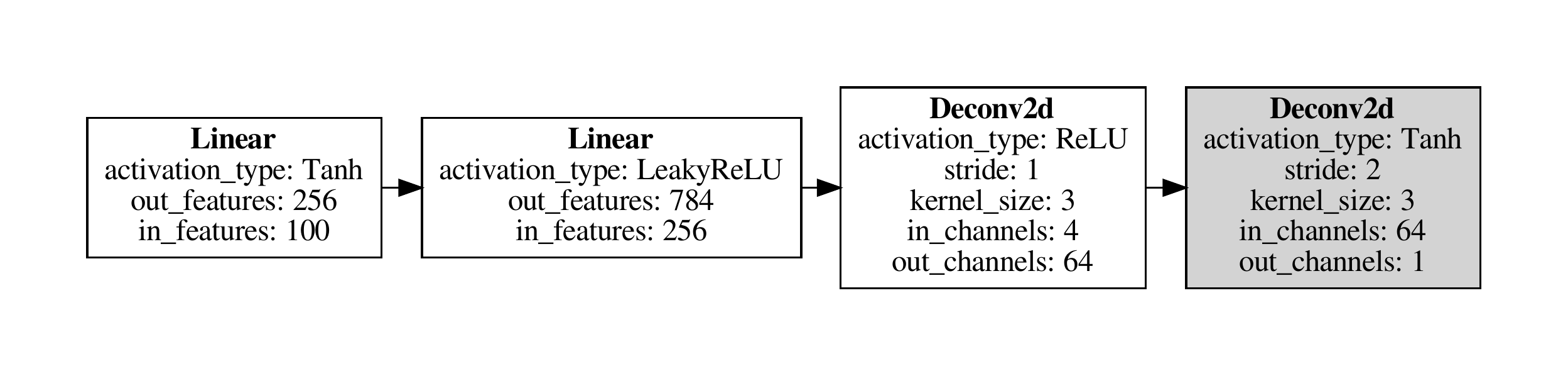}
		\caption{Genotype of a generator}
		\label{fig:mnist_generator}
	\end{subfigure}
	\caption{Best \subref{fig:mnist_discriminator} discriminator and \subref{fig:mnist_generator} generator found after the final generation.}
	\label{fig:mnist_architectures}
\end{figure}

Figure \ref{fig:mnist_architectures} presents the network architecture discovered after the last generation.
We can see that both components reached the limit of four layers imposed in the experiments.
Furthermore, both the generator and the discriminator were composed by a combination of convolutional and linear layers with different activation functions.

\section{Conclusions} \label{sec:conclusions}
Generative adversarial networks (GAN) achieved important results for generative models in the field of computer vision.
However, there are stability problems in the training method, such as the vanishing gradient and the mode collapse problems.

We present in this paper a model that combines neuroevolution and coevolution in the coordination of the GAN training algorithm.
To design the model, we took inspiration on previous evolutionary algorithms, such as NEAT~\cite{stanley2004competitive} and DeepNeat~\cite{miikkulainen2017evolving}, and on recent advances in GANs, such as \cite{karras2018progressive}.

We made experiments using the MNIST dataset.
In all executions, the system reached an equilibrium and convergence, not falling into the vanishing gradient problem.
We also evidenced the FID score as a good fitness metric for generators.
We used the loss function as fitness for the population of discriminators.
However, the results evidenced that a better metric may be necessary to ensure the proper evolution of this population.
The results showed that our model partially solves the mode collapse problem.
In our experiments, the generated samples always presented some diversity, partially preserving the characteristics of the input distribution.
Besides that, our proposal did not outperform the state-of-the-art results, such as presented \cite{lucic2017gans}.

Therefore, as future works, we will experiment other metrics to be used as fitness for the discriminator component, such as area under the curve (AUC).
We will also assess that the proposed model does not suffer from cyclic issues commonly seen in coevolution models \cite{ficici2003game,monroy2006coevolution}.
More experiments should be made with larger populations as well as larger genotype limits.
In addition, the model will also be evaluated with the CelebA dataset \cite{liu2015deep}.
We will make experiments with a reduced growth rate of the networks to avoid premature optimization.
Finally, we will conduct ablation studies to assess the individual contribution of each aspect of the proposed algorithm (e.g., speciation, mutation, crossover).

\section*{Acknowledgments}\label{sec:acknowledgments}
This article is based upon work from COST Action CA15140: ImAppNIO, supported by COST (European Cooperation in Science and Technology): \url{www.cost.eu}.

\bibliographystyle{splncs03}
\bibliography{costa}

\begin{thebibliography}{10}
\providecommand{\url}[1]{\texttt{#1}}
\providecommand{\urlprefix}{URL }

\bibitem{NIPS2014_5423}
Goodfellow, I., Pouget-Abadie, J., Mirza, M., Xu, B., Warde-Farley, D., Ozair,
  S., Courville, A., Bengio, Y.: Generative adversarial nets. In: NIPS (2014)

\bibitem{arjovsky2017wasserstein}
Arjovsky, M., Chintala, S., Bottou, L.: Wasserstein generative adversarial
  networks. In: International Conference on Machine Learning. pp. 214--223
  (2017)

\bibitem{karras2018progressive}
Karras, T., Aila, T., Laine, S., Lehtinen, J.: Progressive growing of {GAN}s
  for improved quality, stability, and variation. In: International Conference
  on Learning Representations (2018)

\bibitem{mao2017least}
Mao, X., Li, Q., Xie, H., Lau, R.Y., Wang, Z., Smolley, S.P.: Least squares
  generative adversarial networks. In: 2017 IEEE International Conference on
  Computer Vision (ICCV). pp. 2813--2821. IEEE (2017)

\bibitem{gulrajani2017improved}
Gulrajani, I., Ahmed, F., Arjovsky, M., Dumoulin, V., Courville, A.C.: Improved
  training of wasserstein gans. In: Advances in Neural Information Processing
  Systems. pp. 5769--5779 (2017)

\bibitem{salimans2016improved}
Salimans, T., Goodfellow, I., Zaremba, W., Cheung, V., Radford, A., Chen, X.:
  Improved techniques for training gans. In: Advances in Neural Information
  Processing Systems. pp. 2234--2242 (2016)

\bibitem{neat}
Stanley, K.O., Miikkulainen, R.: Evolving neural networks through augmenting
  topologies. Evolutionary computation  10(2),  99--127 (2002)

\bibitem{stanley2004competitive}
Stanley, K.O., Miikkulainen, R.: Competitive coevolution through evolutionary
  complexification. Journal of Artificial Intelligence Research  21,  63--100
  (2004)

\bibitem{miikkulainen2017evolving}
Miikkulainen, R., Liang, J., Meyerson, E., Rawal, A., Fink, D., Francon, O.,
  Raju, B., Navruzyan, A., Duffy, N., Hodjat, B.: Evolving deep neural
  networks. arXiv preprint arXiv:1703.00548  (2017)

\bibitem{lecun1998mnist}
LeCun, Y.: The mnist database of handwritten digits. http://yann. lecun.
  com/exdb/mnist/  (1998)

\bibitem{sims1994evolving}
Sims, K.: Evolving 3d morphology and behavior by competition. Artificial life
  1(4),  353--372 (1994)

\bibitem{assunccao2018evolving}
Assun{\c{c}}{\~a}o, F., Louren{\c{c}}o, N., Machado, P., Ribeiro, B.: Evolving
  the topology of large scale deep neural networks. In: European Conference on
  Genetic Programming. pp. 19--34. Springer (2018)

\bibitem{hillis1990co}
Hillis, W.D.: Co-evolving parasites improve simulated evolution as an
  optimization procedure. Physica D: Nonlinear Phenomena  42(1-3),  228--234
  (1990)

\bibitem{rawal2010constructing}
Rawal, A., Rajagopalan, P., Miikkulainen, R.: Constructing competitive and
  cooperative agent behavior using coevolution. In: Computational Intelligence
  and Games (CIG), 2010 IEEE Symposium on. pp. 107--114 (2010)

\bibitem{garcia2003covnet}
Garc{\'\i}a-Pedrajas, N., Herv{\'a}s-Mart{\'\i}nez, C., Mu{\~n}oz-P{\'e}rez,
  J.: Covnet: a cooperative coevolutionary model for evolving artificial neural
  networks. IEEE Transactions on Neural Networks  14(3),  575--596 (2003)

\bibitem{garcia2005cooperative}
Garc{\'\i}a-Pedrajas, N., Herv{\'a}s-Mart{\'\i}nez, C., Ortiz-Boyer, D.:
  Cooperative coevolution of artificial neural network ensembles for pattern
  classification. IEEE transactions on evolutionary computation  9(3),
  271--302 (2005)

\bibitem{gomez2008accelerated}
Gomez, F., Schmidhuber, J., Miikkulainen, R.: Accelerated neural evolution
  through cooperatively coevolved synapses. Journal of Machine Learning
  Research  9,  937--965 (2008)

\bibitem{lucic2017gans}
Lucic, M., Kurach, K., Michalski, M., Gelly, S., Bousquet, O.: Are gans created
  equal? a large-scale study. arXiv preprint arXiv:1711.10337  (2017)

\bibitem{wang2018evolutionary}
Wang, C., Xu, C., Yao, X., Tao, D.: Evolutionary generative adversarial
  networks. arXiv preprint arXiv:1803.00657  (2018)

\bibitem{heusel2017gans}
Heusel, M., Ramsauer, H., Unterthiner, T., Nessler, B., Hochreiter, S.: Gans
  trained by a two time-scale update rule converge to a local nash equilibrium.
  In: Advances in Neural Information Processing Systems. pp. 6629--6640 (2017)

\bibitem{szegedy2016rethinking}
Szegedy, C., Vanhoucke, V., Ioffe, S., Shlens, J., Wojna, Z.: Rethinking the
  inception architecture for computer vision. In: Proceedings of the IEEE
  Conference on Computer Vision and Pattern Recognition. pp. 2818--2826 (2016)

\bibitem{russakovsky2015imagenet}
Russakovsky, O., Deng, J., Su, H., Krause, J., Satheesh, S., Ma, S., Huang, Z.,
  Karpathy, A., Khosla, A., Bernstein, M., et~al.: Imagenet large scale visual
  recognition challenge. International Journal of Computer Vision  115(3),
  211--252 (2015)

\bibitem{tieleman2012lecture}
Tieleman, T., Hinton, G.: Lecture 6.5-rmsprop: Divide the gradient by a running
  average of its recent magnitude. COURSERA: Neural networks for machine
  learning  4(2),  26--31 (2012)

\bibitem{kingma2015adam}
Kingma, D.P., Ba, J.: Adam: A method for stochastic optimization. In:
  International Conference on Learning Representations (ICLR) (2015)

\bibitem{ficici2003game}
Ficici, S.G., Pollack, J.B.: A game-theoretic memory mechanism for coevolution.
  In: Genetic and Evolutionary Computation Conference. pp. 286--297. Springer
  (2003)

\bibitem{monroy2006coevolution}
Monroy, G.A., Stanley, K.O., Miikkulainen, R.: Coevolution of neural networks
  using a layered pareto archive. In: Proceedings of the 8th annual conference
  on Genetic and evolutionary computation. pp. 329--336. ACM (2006)

\bibitem{liu2015deep}
Liu, Z., Luo, P., Wang, X., Tang, X.: Deep learning face attributes in the
  wild. In: Proceedings of the IEEE International Conference on Computer
  Vision. pp. 3730--3738 (2015)

\end{thebibliography}

\end{document}